\documentclass{article}
\usepackage{ijcai21}

\usepackage{times}
\usepackage{soul}
\usepackage{url}
\usepackage[hidelinks]{hyperref}
\usepackage[utf8]{inputenc}
\usepackage[small]{caption}
\usepackage{graphicx}
\usepackage{amsmath}
\usepackage{tabu}
\usepackage{amsthm}
\usepackage{booktabs}
\usepackage{algorithm}
\usepackage{algorithmic}
\usepackage[skip=0pt]{caption}
\usepackage{subfigure}
\usepackage{xspace}
\usepackage[dvipsnames]{xcolor}
\usepackage{multirow}
\usepackage{enumitem}

\newcommand{\paratitle}[1]{\noindent\textbf{#1}}
\newcommand{\ie}{\emph{i.e.,}\xspace}

\newcommand{\eg}{\emph{e.g.,}\xspace}
\newcommand{\etal}{\emph{et al.}\xspace}
\newcommand{\ignore}[1]{}

\newcommand{\citeal}[1]{\citeauthor{#1}~[\citeyear{#1}]}

\newcommand{\tabincell}[2]{\begin{tabular}{@{}#1@{}}#2\end{tabular}}

\newcommand{\jhcomment}[1]{\textcolor{magenta}{[JH: #1]}}
\newcommand{\yscomment}[1]{\textcolor{orange}{[YS: #1]}}

\newcommand{\glcomment}[1]{\textcolor{red}{[GL: #1]}}
\newcommand{\gladd}[1]{\textcolor{red}{#1}}

\title{A Survey on Complex Knowledge Base Question Answering:\\ Methods, Challenges and Solutions}

\author{
Yunshi Lan$^1$\thanks{\ \ Equal contribution.}
\and
Gaole He$^{2,3}$\footnotemark[1] \and
Jinhao Jiang$^{4}$
\and
Jing Jiang$^1$
\and
\\
Wayne Xin Zhao$^{3,4}$\thanks{\ \ Corresponding author.}
\And
Ji-Rong Wen$^{2,3,4}$
\affiliations
$^1$School of Computing and Information Systems, Singapore Management University\\
$^2$School of Information, Renmin University of China\\
$^3$Beijing Key Laboratory of Big Data Management and Analysis Methods\\
$^4$Gaoling School of Artificial Intelligence, Renmin University of China\\
\emails
\{yslan, jingjiang\}@smu.edu.sg,
\{hegaole, jrwen\}@ruc.edu.cn,
\{batmanfly, jiangjinhaonlp\}@gmail.com
}

\begin{document}

\maketitle

\begin{abstract}
  Knowledge base question answering (KBQA) aims to answer a question over a knowledge base (KB).
  Recently, a large number of studies focus on semantically or syntactically complicated questions.
  In this paper, we elaborately summarize the typical challenges and solutions for complex KBQA. 
  We begin with introducing the background about the KBQA task. 
  Next, we present the two mainstream categories of methods for complex KBQA, namely semantic parsing-based~(SP-based) methods and information retrieval-based~(IR-based) methods.
  We then review the advanced methods comprehensively from the perspective of the two categories.
  Specifically, we explicate their solutions to the typical challenges.
  Finally, we conclude and discuss some promising directions for future research.
\end{abstract}

\section{Introduction}
\label{sec:intro}

A knowledge base~(KB) is a structured database that contains a collection of facts in the form (\emph{subject}, \emph{relation}, \emph{object}). %
Large-scale KBs, such as Freebase~\cite{Bollacker-SIGMOD-2008}, DBPedia~\cite{dbpedia-2015} and Wikidata~\cite{Thomas-WWW-2016}, have been constructed to serve many downstream tasks. 
Based on available KBs, knowledge base question answering~(KBQA) is a task that aims to answer natural language questions with KBs as its knowledge source. 
Early work on KBQA~\cite{Bordes-ArXiv-2015,Dong-ACL-2015,Hu-TKDE-2018,Lan-IJCAI-2019,Lan-TASLP-2019} focuses on answering a simple question, where only a single fact is involved. 
For example, ``\textit{Where was JK Rowling born?}'' is a simple question which can be answered using just the fact ``\textit{(J.K. Rowling, birthplace, United Kingdom)}''.

\begin{figure}[t!]
    \centering
    \includegraphics[width=0.48\textwidth]{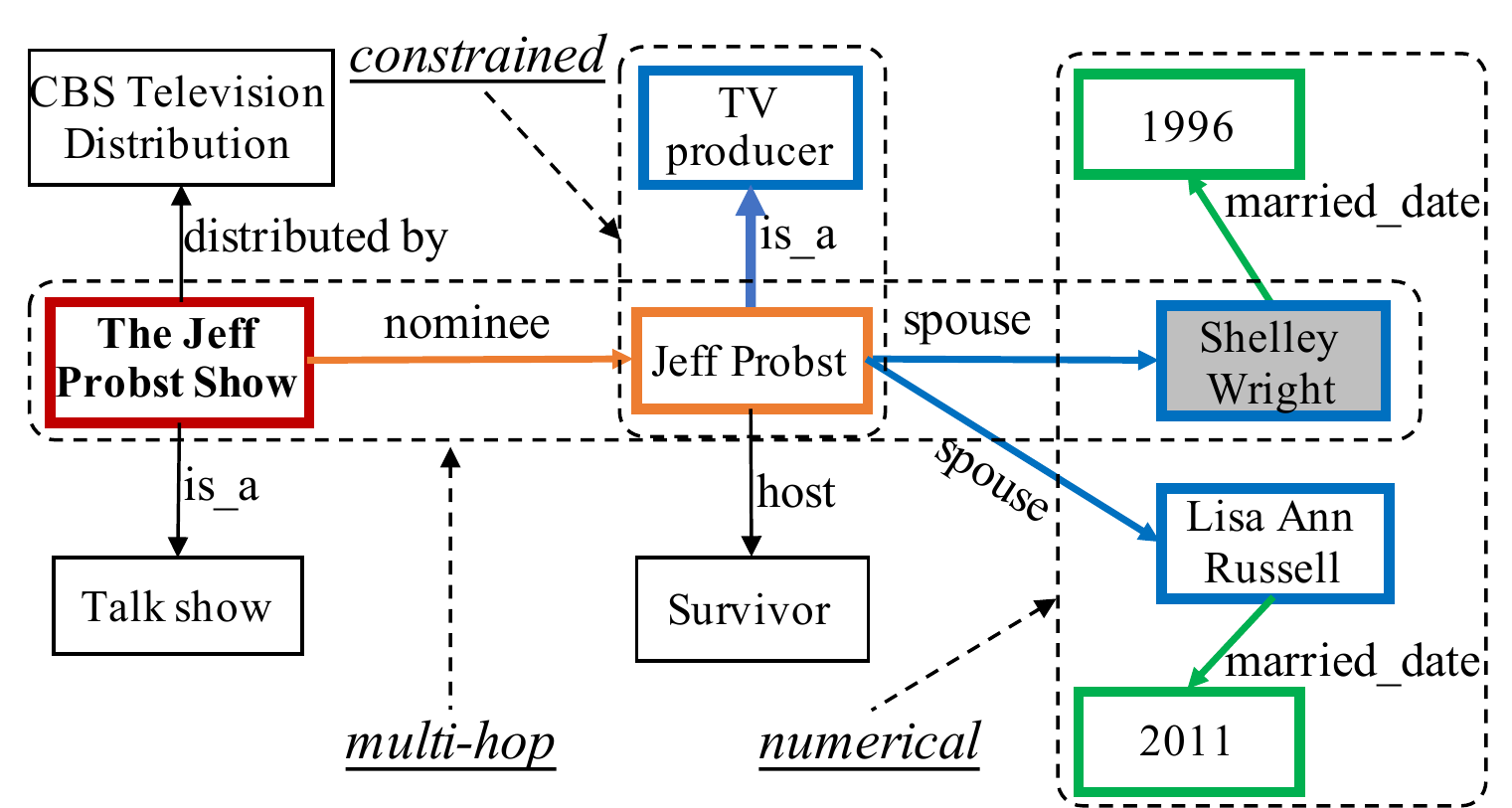}
    \caption{An example of complex KBQA for the question ``\textit{Who is the first wife of TV producer that was nominated for The Jeff Probst Show?}''. 
    We present the related KB subgraph for this question.
    The ground truth path to answer this question is annotated with colored borders. 
    The topic entity and the answer entity are shown in the bold font and shaded box respectively.
    ``multi-hop'' reasoning, ``constrained'' relations and ``numerical'' operation are highlighted in black dotted box. We use different colors to indicate different reasoning hops to reach each entity from the topic entity.
    }
    \label{fig:example}
\end{figure}

Recently, researchers start paying more attention to answering \emph{complex questions} over KBs, i.e., the complex KBQA task~\cite{Hu-EMNLP-18,Luo-EMNLP-2018}.
Complex questions usually contain multiple subjects, express compound relations and include numerical operations.
Take the question in Figure~\ref{fig:example} as an example.
This example question starts with the subject ``\textit{The Jeff Probst Show}''.
Instead of querying a single fact, the question requires the composition of two relations, namely, ``\textit{nominee}'' and ``\textit{spouse}''.
This query is also associated with an entity type constraint ``\textit{(Jeff Probst, is a, TV producer)}''. 
The final answer should be further aggregated by selecting the possible candidates with the earliest marriage date.
Generally, complex questions are questions involving \underline{\emph{multi-hop}} reasoning,  \underline{\emph{constrained}} relations, \underline{\emph{numerical}} operations, or some combination of the above.
Tracing back to the solutions for simple KBQA, a number of studies from two mainstream approaches have been proposed. 
These two approaches first recognize the subject in a question and link it to an entity in the KB (referred to as the \emph{topic entity}).
Then they derive the answers within the neighborhood of the topic entity by either executing a parsed logic form or reasoning in a question-specific graph extracted from the KB.
The two categories of methods are commonly known as \emph{semantic parsing-based methods} (SP-based methods) and \emph{information retrieval-based methods} (IR-based methods) in prior work~\cite{Bordes-ArXiv-2015,Dong-ACL-2015,Hu-TKDE-2018,GrailQA-2020}. 
They include different working mechanisms to solve the KBQA task. 
The former approach represents a question by a symbolic logic form and then executes it against the KB and obtains the final answers. The latter approach constructs a question-specific graph delivering the comprehensive information related to the question and ranks all the entities in the extracted graph based on their relevance to the question.
\ignore{They hold different assumptions.
The former one assumes that a question could always be precisely represented by a symbolic logic form, which is leveraged to execute against KBs and obtain the final answer.
The latter believes: a question-specific graph could expressively deliver the comprehensive information related to the question, and ranking all entities in the graph based on their relevance to the question could predict the answer.
}

However, when applying the two mainstream approaches to the complex KBQA task, complex questions bring in challenges on different parts of the approaches.
We identify the main challenges as follows:

\begin{itemize}
    \item Parsers used in existing SP-based methods are difficult to cover diverse complex queries (\eg multi-hop reasoning, constrained relations and numerical operations). 
    Similarly, previous IR-based methods may fail to answer a complex query, as their ranking is performed over small-scope entities without traceable reasoning.
    \item More relations and subjects in complex questions indicate a larger search space of potential logic forms for parsing, which will dramatically increase  the computational cost.
    Meanwhile, more relations and subjects could prevent IR-based methods from retrieving all relevant entities for ranking.
    \item Both approaches treat question understanding as a primary step. When questions become complicated in both semantic and syntactic aspects, models are required to have strong capabilities of natural language understanding and generalization.
    \item It is expensive to label the ground truth paths to the answers (see the example in Figure~\ref{fig:example}) for complex questions.
    Generally, only question-answer pairs are provided.
    This indicates SP-based methods and IR-based methods have to be trained without the annotation of correct logic forms and reasoning paths, respectively.
    Such weak supervision signals bring difficulties to both approaches.
\end{itemize}

Regarding the related surveys, we observe \citeauthor{Wu-CCK-2019}~[\citeyear{Wu-CCK-2019}] and \citeauthor{Chakraborty-arXiv-2019}~[\citeyear{Chakraborty-arXiv-2019}] reviewed the existing work on simple KBQA. 
Furthermore, \citeauthor{Fu-Arxiv-2020}~[\citeyear{Fu-Arxiv-2020}] investigated the current advances on complex KBQA.
They provided a general view of advanced methods only from the perspective of techniques and focused more on application scenarios in the e-commerce domain.
Different from these surveys, our work tries to identify the challenges encountered in previous studies and extensively discusses existing solutions in a comprehensive and well-organized manner. 
Specifically,  we categorize the methods for complex KBQA into two mainstream approaches based on their working mechanisms.
We decompose the overall procedure of the two approaches into a series of modules and analyze the challenges in each module. 
We believe that this way is particularly helpful for readers to understand the challenges and how they are addressed in existing solutions to complex KBQA.
Furthermore, we provide a thorough outlook on several promising research directions on complex KBQA.



\ignore{This paper is organized as follows.
First, we introduce multiple available datasets for conducting complex KBQA researches.
Second, we summarize and compare the advanced approaches.
Finally, we conclude and discuss future research directions.}

The remainder of this survey is organized as follows.
We will first introduce the preliminary knowledge about the task formulation,  multiple available datasets and evaluation protocol in Section~\ref{sec:backgroung}.
Next, we introduce the two mainstream categories of methods for complex KBQA in Section~\ref{sec-methods}. 
Then following the categorization, we figure out typical challenges and solutions to these challenges in Section~\ref{sec:challenge_solution}. 
Finally, we conclude and discuss some future research directions in Section~\ref{sec:con}.









\begin{table}[!tbp]
	\centering
	\begin{scriptsize}
	\begin{tabular}{c | c c c c}
		\hline
		\multirow{2}{*}{Datasets}&	\multirow{2}{*}{KB}&	\multirow{2}{*}{Size}&	\multirow{2}{*}{LF}& \multirow{2}{*}{NL}\\
			&	&	&	&	\\
		\hline
		\hline
		WebQuestions~\cite{Berant-EMNLP-2013}&	Freebase&	5,810&	No&	No\\
		ComplexQuestions~\cite{Bao-COLING-2016}&	Freebase&	2,100&	No&	No\\
		WebQuestionsSP~\cite{Yih-ACL-2016}&	Freebase&	4,737&	Yes&	Yes\\
        ComplexWebQuestions&	\multirow{2}{*}{Freebase}&	\multirow{2}{*}{34,689}&	\multirow{2}{*}{Yes}&	\multirow{2}{*}{Yes}\\
		\cite{CWQ-NAACL-2018}&	&	&	&\\
		\hline
		QALD series~\cite{QALD}&	DBpedia&	-&	Yes&	Yes\\
		LC-QuAD~\cite{LC-QuAD-ISWC-2017}&	DBpedia&	5,000&	Yes&	Yes\\
		\multirow{2}{*}{LC-QuAD 2.0~\cite{LC-QuAD-2.0-ISWC-2019}}&	\multirow{2}{*}{\tabincell{c}{DBpedia,\\Wikidata}}&	\multirow{2}{*}{30,000}&	\multirow{2}{*}{Yes}&	\multirow{2}{*}{Yes}\\
		&	&	&	&\\
		\hline
		MetaQA	Vanilla~\cite{Zhang-AAAI-2018}&	WikiMovies&	400k&	No&	No\\
		CFQ~\cite{CFQ-ICLR-2020}& Freebase&	239,357&	Yes&	No\\
		GrailQA~\cite{GrailQA-2020}&	Freebase&	64,331&	Yes&Yes\\
		KQA Pro~\cite{KQA-Pro}&	Wikidata&	117,970&	Yes&	Yes\\
		\hline
	\end{tabular}%
	\caption{Several complex KBQA benchmark datasets. ``\textbf{LF}" denotes whether the dataset provides \textbf{L}ogic \textbf{F}orms, and ``\textbf{NL}" denotes whether the dataset incorporates crowd workers to rewrite questions in \textbf{N}atural \textbf{L}anguage.
	}
	\label{tab:datasets}%
	\end{scriptsize}
\end{table}%

\section{Background}
\label{sec:backgroung}
In this section, we first give a task definition about complex KBQA, and then introduce available  datasets and  evaluation protocol for this task.

\begin{itemize}[leftmargin=0cm, label={}]
  \item \textbf{Task}. For the task of complex KBQA, a    KB consisting of a set of facts is given as input,    where the subject and object are connected by their   relation. 
    All the subjects and objects in the facts form the entity set of a KB.
    Given the available  KB, this task aims to answer complex natural language questions in the format of a sequence of tokens. 
    Specially, we assume the correct answers come from the entity set of the KB. 
    Unlike answers of simple KBQA, which are entities directly connected to the topic entity, the answers of the complex KBQA task are entities multiple hops away from the topic entities or even some aggregation of them.
  \item \textbf{Datasets}. Generally, the answers of the questions should be provided to train a complex KBQA system. For this purpose, many efforts have been devoted to constructing datasets for complex KBQA. We list the available  complex KBQA datasets in Table~\ref{tab:datasets}.
  Overall, these datasets are constructed with the following steps. Given a topic entity in a KB as question subject, simple questions are first created with diverse templates. Based on simple questions and the neighborhood of a topic entity in a KB, complex questions are further generated with predefined templates, and some work~\cite{KQA-Pro} also generates executable logic forms with templates. Meanwhile, answers are extracted with corresponding rules. In some cases, crowd workers are hired to paraphrase the template queries into natural language questions and refine the generated logic forms, making the question expressions more diverse and fluent. In order to serve realistic applications, these datasets typically create questions which require multiple KB facts to reason. Moreover, they might include numerical operations (\eg counting, ranking operations for comparative or superlative questions) and constraints (\eg entity, temporal keywords), which further increase the difficulty in reasoning the answers from KBs.
  \item \textbf{Evaluation Protocol}. 
The KBQA system usually predicts entities with the top confidence score to form the answer set.
Note that there can be more than one answer to a question.
In previous studies, there are some classical evaluation metrics such as \textit{precision}, \textit{recall}, $F_1$ and \textit{Hits@1}.
Some studies~\cite{Yih-ACL-2015,Liang-ACL-17,Abujabal-WWW-17} use the \textit{precision}, \textit{recall}, $F_1$ score to evaluate the prediction. 
\textit{Precision} indicates the ratio of the correct answers over all the predicted  answers. 
\textit{Recall} is the ratio of the correct predicted answers over all the ground truth. 
And $F_1$ score considers \textit{precision} and \textit{recall} simultaneously.
Some other methods~\cite{Miller-EMNLP-2016,Sun-EMNLP-2018,Xiong-ACL-2019,He-WSDM-2021} use \textit{Hits@1} to assess the fraction that the correct answers rank higher than other entities.
\end{itemize}


\begin{figure}[!t]
 \centering
 \includegraphics[width=0.48\textwidth]{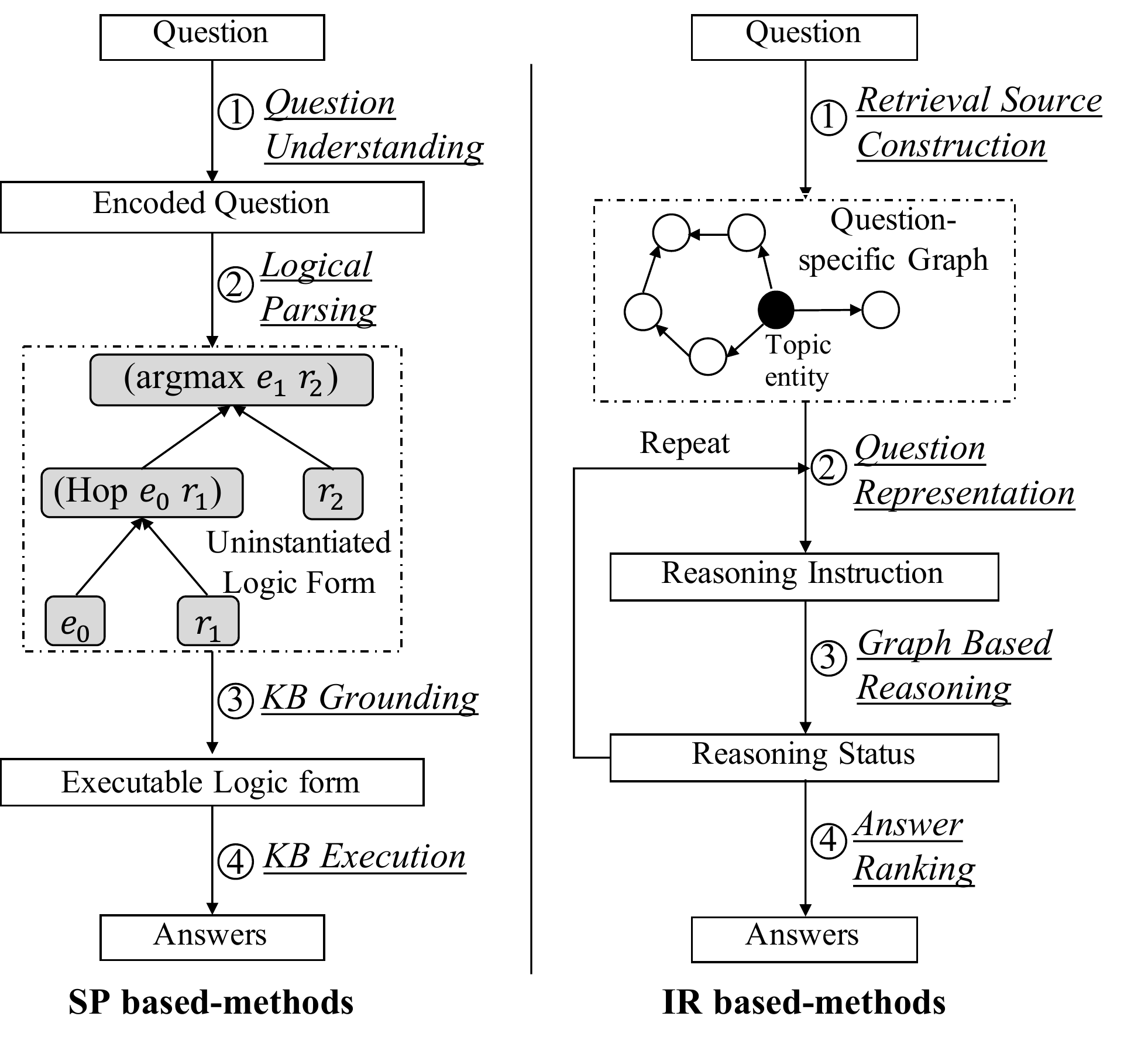}
 \centering
 \caption{Illustration of two mainstream approaches for complex KBQA.
 }
 \label{fig-methods}
\end{figure}

\section {Two Mainstream Approaches}
\label{sec-methods}

As introduced in Section~\ref{sec:intro}, SP-based and IR-based methods are two mainstream approaches to solving complex KBQA task.
SP-based methods parse a question into a logic form and execute it against KBs for finding the answers.
IR-based methods retrieve a question-specific graph and apply some ranking algorithms to select entities from top positions.
To summarize, the two approaches follow either a \emph{parse-then-execute} paradigm or 
a \emph{retrieval-and-rank} paradigm, which are illustrated in Figure~\ref{fig-methods}. 


\begin{itemize}[leftmargin=0cm, label={}]
\item \textbf{Semantic Parsing-based Methods}.  
This category of methods aims at parsing a natural language utterance into a logic form~\cite{Berant-ACL-2014,Reddy-TACL-2014}.
They predict answers via the following steps:
(1) They fully understand a question via a \underline{\emph{question understanding}} module, which is to conduct the semantic and syntactic analysis and obtain an encoded question for the subsequent parsing step.
(2) A \underline{\emph{logical parsing}} module is utilized to transfer the encoded question into an uninstantiated logic form.
The uninstantiated logic form is a syntactic representation of the question without the grounding of entities and relations.
The grammar and constituents of logic forms could be different according to specific designs of a system.
(3) To execute against KBs, the logic form is further instantiated and validated by conducting some semantic alignments to structured KBs via \underline{\emph{KB grounding}}. 
Note that, in some work~\cite{Yih-ACL-2015,Liang-ACL-17}, the logical parsing and KB grounding are simultaneously performed, where logic forms are validated in KBs while partially parsed.
(4) Eventually, the parsed logic form is executed against KBs to generate predicted answers via a \underline{\emph{KB execution}} module.
\item \textbf{Information Retrieval-based Methods}. 
As another mainstream approach, IR-based methods  directly retrieve and rank answers from the KBs considering the information conveyed in the questions~\cite{Bordes-ArXiv-2015,Dong-ACL-2015}.
They consist of the following steps:
(1) Starting from the topic entity, the system first extracts a question-specific graph from KBs. 
Ideally, this graph includes all question related entities and relations as nodes and edges, respectively.
Without explicitly generating an executable logic form, IR-based methods perform reasoning over the graph and then rank entities in the graph. 
(2) Next, the system encodes input questions via a \underline{\emph{question representation}} module.
This module analyzes the semantics of the question and outputs reasoning instructions, which are usually represented as vectors.
(3) A \underline{\emph{graph-based reasoning}} module conducts semantic matching via vector-based computation to propagate and then aggregate the information along the neighboring entities within the graph.
The reasoning status, which has diverse definitions in different methods (\eg distributions of predicted entities, representations of relations), is updated based on the reasoning instruction. 
Recently, several studies~\cite{Jain-ACL-2016,Chen-NAACL-2019} repeat Step (2) and (3) for multiple times to perform the reasoning.
(4) An \underline{\emph{answer ranking}}  module is utilized to rank the entities in the graph according to the reasoning status at the end of reasoning. 
The top-ranked entities are predicted as the answers to the question.
\item \textbf{Pros and Cons}. Overall, SP-based methods can produce a more interpretable reasoning process by generating expressive logic forms. 
However, they heavily rely on the design of the logic form and parsing algorithm, which turns out to be the bottleneck of performance improvement. As a comparison, IR-based methods conduct complex reasoning on graph structure and perform semantic matching. Such a paradigm naturally fits into popular end-to-end training and makes the IR-based methods easier to train. 
However, the blackbox style of the reasoning model makes the intermediate reasoning less interpretable.
\end{itemize}

\section{Challenges and Solutions}
\label{sec:challenge_solution}

\begin{table*}[tbp]
	\centering
	\begin{footnotesize}
	\begin{tabular}{p{0.07\textwidth}| p{0.11\textwidth} |p{0.13\textwidth} |p{0.59\textwidth}}
		\hline
		Categories & Modules & Challenges & Solutions  \\
		\hline \hline
		\multirow{13}{*}{\tabincell{c}{SP-based\\ Methods}} & Question understanding & Understanding complex semantics and syntax & Adopt structural properties (\eg dependency parsing~\cite{Abujabal-WWW-17,Abujabal-WWW-2018,Luo-EMNLP-2018},  AMR~\cite{Kapanipathi-AAAI-2021}) augmented parsing, skeleton-based parsing ~\cite{Skeleton-AAAI-2020} or structural properties based matching~\cite{Maheshwari-ISWC-2019,Zhu-Neuro-2020,Chen-IJCAI-2020}.\\
		\cline{2-4}
		& Logical parsing & Parsing complex queries & Develop expressive targets for parsing, such as: template based queries~\cite{Bast-CIKM-15}, query graph~\cite{Yih-ACL-2015,Abujabal-WWW-17,Hu-EMNLP-18}, and so on.  \\
		\cline{2-4}
		& KB grounding & Grounding with large search space & Narrow down search space by decompose-execute-join strategy~\cite{Zheng-VLDB-2018,Bhutani-CIKM-2019} or expand-and-rank strategy~\cite{Chen-NAACL-2019,Lan-ICDM-2019,Lan-ACL-2020}. \\
		\cline{2-4}
		& Training procedure & Training under weak supervision signals & Adopt reward shaping strategy to strengthen training signal~\cite{Saha-TACL-2019,Hua-JWS-2020,Qiu-CIKM-2020}, conduct pre-training to initialize the model~\cite{Qiu-CIKM-2020} or iterative maximum-likelihood training\cite{Liang-ACL-17}. \\
		\hline
		\multirow{11}{*}{\tabincell{c}{IR-based\\Methods}}& Retrieval~source construction & Reasoning under incomplete KB & Supplement KB with extra corpus~\cite{Sun-EMNLP-2018,Sun-EMNLP-2019}, fuse extra textual information into entity representations~\cite{Xiong-ACL-2019,Han-EMNLP-2020} or leverage KB embeddings~\cite{Apoorv-ACL-2020}.\\
		\cline{2-4}
		& Question representation & Understanding~complex semantics & 
		Update with reasoned information~\cite{Miller-EMNLP-2016,Zhou-COLING-2018,Xu-NAACL-2019}, dynamic attention over the question~\cite{He-WSDM-2021} or enrich the question representation with contextual information of  graph~\cite{Sun-EMNLP-2018}. \\
		\cline{2-4}
		& Graph based reasoning & Uninterpretable~reasoning& Provide traceable reasoning path~\cite{Zhou-COLING-2018,Xu-NAACL-2019} or  hyperedge based reasoning~\cite{Han-IJCAI-2020}.\\
		\cline{2-4}
		& Training procedure & Training under weak supervision signals & Provide shaped reward as intermediate feedback~\cite{Qiu-WSDM-2020},  augment intermediate supervision signals with bidirectional search algorithm~\cite{He-WSDM-2021} or adopt variational algorithm to train entity linking module~\cite{Zhang-AAAI-2018}. \\
		\hline
    \end{tabular}
	\end{footnotesize}
	\caption{Summary of the existing studies on complex KBQA.
    We categorize them into two mainstream approaches \textit{w.r.t.}  key modules and solutions according to different challenges.}
	\label{tab:methods}%
\end{table*}%

Since the aforementioned approaches are developed based on different paradigms, we describe the challenges and corresponding solutions for complex KBQA with respect to the two mainstream approaches. 
A summary of these challenges and solutions is presented in Table~\ref{tab:methods}.

\subsection{Semantic Parsing-based Methods}
In this part, we  discuss the challenges and solutions for semantic parsing-based methods.


\begin{itemize}[leftmargin=0cm, label={}]
\item \textbf{Overview}. 
As introduced in Section~\ref{sec-methods}, SP-based methods follow a parse-then-execute procedure via a series of modules, namely question understanding, logical parsing, KB grounding and KB  execution. These modules will encounter different challenges for complex KBQA.
Firstly, question understanding becomes more difficult when the questions are complicated in both semantic and syntactic aspects.
Secondly, logical parsing has to cover diverse query types of complex questions. 
Moreover, a complex question involving more relations and subjects will dramatically increase the possible search space for parsing, which makes the parsing less effective. 
Thirdly, the manual annotation of logic forms are both expensive and labor-intensive, and it is challenging to train a SP-based method with weak supervision signals (\ie question-answer pairs). 
Next, we will introduce how prior studies deal with these challenges.

\ignore{
To understand complex question and obtain right intention, models are supposed to tackle with \textbf{complex semantics and syntax}. 
After that, model transform the question into logical forms which are capable to accurately express \textbf{complicated  queries}. 
Then, the logical form should be grounded with provided KB, which may contain \textbf{large search space}, and obtain executable logical from (\eg SPARQL). 
Finally, the model is optimized according to the predicted logical form and ground truth answers (definitely, only such \textbf{weak supervision signals} are provided). 
}

\item \textbf{Understanding Complex Semantics and Syntax}. 
As the first step of SP-based methods, question understanding module converts unstructured text into encoded question (\ie structural representation), which benefits the downstream parsing. Compared with simple questions, complex questions are featured with more complex query types and compositional semantics, which increases the difficulty in linguistic analysis.
To better understand complex natural language questions, many existing methods rely on syntactic parsing, such as dependencies~\cite{Abujabal-WWW-17,Abujabal-WWW-2018,Luo-EMNLP-2018} and Abstract Meaning Representation~(AMR)~\cite{Kapanipathi-AAAI-2021}, to provide better alignment between question constituents and logic form elements (\eg entity, relation, entity types and attributes). 
However, the accuracy of producing syntactic parsing is still not  satisfying on complex questions, especially for those with long-distance dependency. 
To alleviate error propagation from syntactic parsing to downstream semantic parsing, 
\citeal{Skeleton-AAAI-2020} leveraged the skeleton-based parsing to obtain the trunk of a complex question, which is a simple question with several branches (\ie pivot word of original text-spans) to be expanded.
Another line of work focuses on leveraging structural properties (such as tree structure or graph structure) of logic forms for ranking candidate  parsing. 
They try to improve the matching between logic forms and questions by incorporating  structure-aware feature encoder~\cite{Zhu-Neuro-2020}, applying fine-grained slot matching~\cite{Maheshwari-ISWC-2019}, and adding constraints about query structure to filter noisy queries out~\cite{Chen-IJCAI-2020}.

\item \textbf{Parsing Complex Queries}.
During parsing, traditional semantic parses (\eg CCG~\cite{Cai-ACL-2013,Kwiatkowski-EMNLP-2013,Reddy-TACL-2014}), which are developed without considering KB schemas, have shown their potential  in parsing simple questions. However, these methods could be sub-optimal for complex questions due to the ontology mismatching problem~\cite{Yih-ACL-2015}. 
Thus, it is necessary to leverage the structure of KBs for more accurate parsing.
To satisfy the compositionality of the complex questions, researchers have developed diverse expressive logic forms as parsing targets.
\citeauthor{Bast-CIKM-15}~[\citeyear{Bast-CIKM-15}] designed three query templates as the parsing targets, which could cover questions querying 1-hop, 2-hop relations and single constraint involved relations.
Although this piece of work can successfully parse several types of complex questions, it suffers from the limited coverage issue.
\citeauthor{Yih-ACL-2015}~[\citeyear{Yih-ACL-2015}] proposed \textit{query graph} as the expressive parsing target. 
A query graph is a logic form in graph structure which closely matches with the KB schemas.
Such query graphs have shown strong expression capability in complex KBQA task. 
However, they are restrictedly generated with predefined manual rules, which is inapplicable to large-scale datasets and long-tail complex question types. 
The follow-up work tried to improve the formulation of query graphs. 
To generalize to unseen and long-tail question types, \citeauthor{Ding-EMNLP-2019}~[\citeyear{Ding-EMNLP-2019}] proposed to leverage frequent query substructure for formal query generation.
\citeauthor{Abujabal-WWW-17}~[\citeyear{Abujabal-WWW-17}] utilized syntactic annotation to enhance the structural complexity of the query graph.
\citeauthor{Hu-EMNLP-18}~[\citeyear{Hu-EMNLP-18}] applied more aggregation operators (\eg  ``merging'') to fit complex questions, and conducted coreference resolution.

\item \textbf{Grounding with Large Search Space}.
To obtain executable logic forms, KB grounding module instantiates possible logic forms with a KB. As one entity in the KB could be linked to hundreds or even thousands of relations, it would be unaffordable to explore and ground all the possible logic forms for a complex question considering both computational resource and time complexity. 
Recently, researchers proposed multiple approaches to solving the problem.
\citeauthor{Zheng-VLDB-2018}~[\citeyear{Zheng-VLDB-2018}]  proposed to decompose a complex question into multiple simple questions, where each question was parsed into a simple logic form.
Next, intermediate answers are generated via these simple logic forms and final answers are jointly obtained. 
This \textit{decompose-execute-join} strategy could effectively narrow down the search space.
A similar approach was studied by~\citeal{Bhutani-CIKM-2019} and they reduced human annotations by leveraging dependency structure. 
Meanwhile, a number of studies adopted the \textit{expand-and-rank} strategies to reduce the search space by searching the logic forms with beams.
\citeal{Chen-NAACL-2019} first adopted the hopwise greedy search strategy to expand the most likely query graphs and stop until the best query graph was obtained. 
\citeal{Lan-ICDM-2019} proposed an iterative matching module to parse the questions  without revisiting the generated query graphs at each searching step.
Such a sequential expansion process is only effective in answering multi-hop questions, while helpless for questions with constraints or numerical operations. 
\citeal{Lan-ACL-2020} defined more operations to support three typical complex queries, which can largely reduce the search space.


\item \textbf{Training under Weak Supervision Signals}.
To deal with the issue of limited or insufficient training data, 
 Reinforcement Learning (RL) based optimization has been adopted to maximize the expected reward~\cite{Liang-ACL-17,Qiu-CIKM-2020}. 
In such a way, SP-based methods can only receive the feedback after the execution of the complete parsed logical form, which leads to severe sparse positive rewards and data inefficiency issues. 
To tackle these issues, some research work adopted \emph{reward shaping} strategies for parsing evaluation. \citeal{Saha-TACL-2019} rewarded the model by the additional feedback when the predicted answers are the same type as the ground truth. 
\citeal{Hua-JWS-2020} adopted a similar idea to evaluate the generated logic form by comparing it with the high-reward logic forms stored in the memory buffer. Besides rewards for the whole procedure, intermediate rewards during the semantic parsing process may also help address this challenge. Recently, \citeal{Qiu-CIKM-2020} formulated query graph generation as a hierarchical decision problem, and proposed a framework based on hierarchical RL with intrinsic motivations to provide intermediate rewards.
To accelerate and stablize the training process, Qiu \etal~\cite{Qiu-CIKM-2020} \emph{pre-trained} model with pseudo-gold programs (\ie high-reward logic forms generated by hand-crafted rules). 
As pseudo-gold programs can be also generated from the model,  \citeal{Liang-ACL-17} proposed to maintain pseudo-gold programs found by an iterative maximum-likelihood training process to bootstrap training. 

\end{itemize}

\subsection{Information Retrieval-based Methods}
\ignore{
As mentioned in Sec.~\ref{sec-methods}, information retrieval-based methods generally perform multi-hop reasoning over the question-specific graph to find the answer.
Information retrieval-based methods first construct the question-specific graph as either the entire KBs or the partial subgraph, which face the \textbf{incomplete knowledge} owing to the incompleteness of KBs.
And then, they need to perform multi-hop reasoning over the question-specific graph.
In order to analyse and optimize the answer inference for complex KBQA, the intermediate reasoning status is supposed to be explicitly visiable.
However, existing methods directly viewed the whole reasoning process as vector similarity matching, which caused the \textbf{lack of interpretability}.
Similar as the semantic parsing-based methods, this category also encounters the \textbf{weak supervision signal} when training, owing to the lack of ground truth annotation of multi-hop reasoning path.
The following part illustrate how prior work deal with these challenges.
}
Here, we summarize the main challenges brought by complex questions for different modules of IR-based methods.

\begin{itemize}[leftmargin=0cm, label={}]
\item \textbf{Overview}.  
The overall procedure typically consists of the modules of retrieval source construction, question representation, graph based reasoning and answer ranking.  
These modules will encounter different challenges for complex KBQA. 
Firstly, the retrieval source construction module extracts a question-specific graph from KBs, which covers a wide range of relevant facts for each question. Due to unneglectable incompleteness of source KBs~\cite{Min-NAACL-2013}, the correct reasoning paths may be absent from the extracted graph. This issue is more likely to occur in the case of complex questions. 
Secondly, question representation module understands the question and generates instructions to guide the reasoning process. 
This step becomes challenging when the question is complicated. 
After that, reasoning on graph is conducted through semantic matching.
When dealing with complex questions, such methods rank answers through semantic similarity without traceable reasoning in the graph, which hinders reasoning analysis and failure diagnosis.
Eventually, this system encounters the same training challenge under weak supervision signals (\ie question-answer pairs). 
The following parts illustrate how prior work deal with these challenges.

\item \textbf{Reasoning under Incomplete KB}.
IR-based methods first extract a question-specific graph from KBs, and conduct subsequent reasoning on it. 
Since simple questions only require 1-hop reasoning on the neighborhood of topic entity in KBs, IR-based methods are less likely to suffer from the inherent incompleteness of KBs~\cite{Min-NAACL-2013}. 
In comparison, it may be a severe problem for complex questions, where the correct reasoning path may be absent from the question-specific graph. 
Furthermore, this incompleteness reduces the neighborhood information used for encoding entities, which poses additional challenges for effective  reasoning.
To tackle this challenge, researchers utilize auxiliary information to enrich the knowledge source. 
Intuitively, large question-related text corpus retrieved from Wikipedia can provide a wide range of unstructured knowledge as supplementary evidence.
\citeal{Sun-EMNLP-2018} and \citeal{Sun-EMNLP-2019} proposed to complement the subgraph extracted from incomplete KBs with extra question-related text sentences to form a heterogeneous graph and conduct reasoning on it.
Instead of directly complementing sentences to question-specific graph as nodes, \citeal{Xiong-ACL-2019} and \citeal{Han-EMNLP-2020} proposed to fuse extra textual information into the entity representation to supplement knowledge.
They first encoded sentences and entities conditioned on questions, and then supplemented the incomplete KB by  aggregating representations of sentences to enhance corresponding entity representations.
Besides extra text corpus, knowledge base embeddings have been adopted to alleviate the sparsity of KB by performing missing link prediction.
Inspired by KB completion task, \citeal{Apoorv-ACL-2020} utilized  pre-trained knowledge base embeddings to enrich the learned entity representations and address incomplete KB issue.

\item \textbf{Understanding Complex Semantics}. 
In general, IR-based methods generate reasoning instructions by directly encoding questions as low-dimensional vectors through neural network (\eg LSTM).
Static reasoning instructions obtained through above approaches cannot effectively represent the compositional semantics of complex questions. 
In order to comprehensively understand questions, recent work dynamically updated the reasoning instructions during the reasoning process.
To focus on the currently unanalyzed part of question, \citeal{Miller-EMNLP-2016}, \citeal{Zhou-COLING-2018} and ~\citeal{Xu-NAACL-2019} proposed to update the reasoning instruction with information retrieved along the reasoning process. 
Besides updating the instruction representation with the reasoned information, \citeal{He-WSDM-2021} proposed to focus on different parts of the question with dynamic attention mechanism.
This dynamic attention mechanism can promote the model to attend to other information conveyed by the question  and provide proper guidance for subsequent  reasoning steps. 
Instead of decomposing the semantics of questions, \citeal{Sun-EMNLP-2018} proposed to augment the representation of the question with contextual information from graph.
They updated the reasoning instruction through aggregating information from the topic entity after every reasoning step.

\item \textbf{Uninterpretable Reasoning}.
Traditional IR-based methods rank answers by calculating a single semantic similarity between questions and entities in the graph, which is less interpretable at the intermediate steps.  
As the complex questions usually query multiple facts, the system is supposed to accurately predict answers over the graph based on a traceable and observable reasoning process.
Even though some work repeated reasoning steps for multiple times, they cannot reason along a traceable path in the graph.
To derive a more interpretable reasoning process, multi-hop reasoning is introduced. Specifically, \citeal{Zhou-COLING-2018} and ~\citeal{Xu-NAACL-2019} proposed to make the relation or entity predicted at each hop traceable and observable. 
They output intermediate predictions (\ie matched relations or entities) from predefined memory as the interpretable reasoning path.
Nevertheless, it can not fully utilize the semantic relation information to reason edge by edge. 
Thus, \citeal{Han-IJCAI-2020} constructed a denser hypergraph by pinpointing a group of entities connected via same relation, which simulated human’s hopwise relational reasoning and output a sequential relation path to make the reasoning interpretable.
\item \textbf{Training under Weak Supervision Signals}. 
Similar to the SP-based methods, it is difficult for IR-based methods to reason the correct answers without any annotations at intermediate steps, since the model cannot receive any feedback until the end of reasoning. 
It is found that this case may lead to spurious reasoning~\cite{He-WSDM-2021}.
To mitigate such issues, \citeal{Qiu-WSDM-2020} formulated the reasoning process over KBs as expanding the reasoning path on KB and adopted reward shaping strategy to provide intermediate rewards. 
To evaluate reasoning paths at intermediate steps, they utilized semantic similarity between the question and the reasoning path to provide feedback.
Besides evaluating the reasoning path at intermediate steps, a more intuitive idea is to infer pseudo intermediate status and augment model training with such inferred signals. 
Inspired by bidirectional search algorithm on graph, \citeal{He-WSDM-2021} proposed to learn the intermediate reasoning entity distributions by synchronizing  bidirectional reasoning process.
While most of existing work focused on enhancing the supervision signals at intermediate steps, few work paid attention to the entity linking step.
Researchers utilized off-the-shelf tools to locate the topic entity in question, which may cause error propagation to subsequent reasoning. 
In order to accurately locate the topic entity without annotations, \citeal{Zhang-AAAI-2018} proposed to train entity linking module through a variational learning algorithm which jointly modeled topic entity recognition and subsequent reasoning over KBs.

\end{itemize}

\section{Conclusion and Future Directions}
\label{sec:con}
\ignore{
This paper attempted to provide an overview of recent developments on complex KBQA. 
Obviously, it cannot cover all the literatures on complex KBQA, and we focused on a representative subset of the typical challenges and advanced solutions.
}
This paper attempted to provide an overview of typical challenges and corresponding solutions on complex KBQA. 
We introduced commonly used datasets and summarized the widely employed SP-based methods as well as IR-based methods. 
Existing complex KBQA methods are generally summarized into these two categories. 
Besides them, some other methods~\cite{CWQ-NAACL-2018} may not fall into these two categories. For example, ~\citeal{CWQ-NAACL-2018} proposed to transform a complex question to a composition of simple questions through rule-based decomposition, which focused on question decomposition instead of KB based reasoning or logic form generation. 
We believe that complex KBQA will continue to be an active and promising research area with wide applications, such as natural language understanding, compositional generalization, multi-hop reasoning. Many challenges presented in this survey are still open and under-explored.

\ignore{
\glcomment{Here, we show the setup for future directions, but this list should be deleted later.}
\begin{itemize}
	\item Develop Expressive Targets for Parsing, Narrow Down the Search Space $\rightarrow$ Improved reasoning pipeline.
	\item Train under Weak Supervision, Promote Interpretability of Reasoning Process, Reason under Weak Supervision. $\rightarrow$ Robust and interpretable model.
	\item Augment Complex Parsing with Structural Properties. $\rightarrow$ Understand complex question with syntactic knowledge.
	\item Supplement Incomplete Knowledge Bases. $\rightarrow$ More general grounded knowledge base.
\end{itemize}
}

Considering the challenges summarized in this paper, we point out several promising future directions for complex KBQA task:

\ignore{
\paratitle{Improve reasoning pipeline.}~\yscomment{This direction is too general} 
As we can see that, both semantic parsing-based methods and information retrieval-based methods seek for better training targets, reasoning interpretability, inference accuracy via different techniques, such as incorporating extra information, designing efficient operations, constraining intermediate reasoning process. 
Following these motivations, improving whole complex KBQA pipeline (\eg entity linking, target setting) will keep playing an important role in this research topic.
}

\begin{itemize}[leftmargin=0cm, label={}]
\item \textbf{Evolutionary KBQA}. 
As we can see, existing methods for complex KBQA task are usually learned on offline training datasets and then deployed online to answer user questions. 
Due to such clear separation, most of existing KBQA systems fail to catch up with the rapid growth of world knowledge and answer new questions. 
However, user feedback may provide deployed KBQA systems an opportunity to improve themselves. 
Based on this observation, \citeauthor{Abujabal-WWW-2018}~[\citeyear{Abujabal-WWW-2018}] leveraged the user feedback to rectify answers generated by the KBQA system and made further improvement.
Despite verifying the correctness of system prediction, 
users may also play a more active role in the question answering process.  \citeauthor{Zheng-InfoScience-2018}~[\citeyear{Zheng-InfoScience-2018}] designed an interactive method to engage users in the question parsing process of the KBQA system directly. 
In the future, an evolutionary KBQA system is imperative to get continuous improvement after online deployment. 

\item \textbf{Robust and Interpretable Models}.  
While existing methods have achieved promising results on benchmark datasets where \textit{i.i.d} assumption is held, they may easily fail to deal with an out-of-distribution case. 
Few-shot setting is a scenario where the training data is limited. 
A few previous studies~\cite{Hua-IJCAI-2020,He-WSDM-2021} discussed related topics, but they are still far from comprehensive in terms of challenge anslysis and problem solving. 
Compositional generalization is another scenario where the novel combinations of component items seen in training should be inferred during testing. 
To support more research on such issue, \citeauthor{GrailQA-2020}~[\citeyear{GrailQA-2020}] and \citeauthor{CFQ-ICLR-2020}~[\citeyear{CFQ-ICLR-2020}] have introduced related datasets, namely GrailQA and CFQ. 
The models are supposed to handle out-of-distribution questions and obtain explainable reasoning process. 
Designing methods for KBQA with good interpretability and robustness may be a challenging but promising topic for future research. 

\item \textbf{More General Knowledge Base}. 
Due to KB incompleteness, researchers incorporated extra information (such as text~\cite{Sun-EMNLP-2018}, images~\cite{Xie-IJCAI-2017} and human interactions~\cite{He-WWW-2020}) to complement the knowledge base, which would further improve the complex KBQA performance. There are also some tasks (\eg visual question answering and commonsense knowledge reasoning), which can be formulated as question answering based on specific KBs. 
For example, in visual question answering, the scene graph extracted from an image can be regarded as a special KB~\cite{Hudson-NeurIPS-2019}. 
Despite explicitly representing relational knowledge as the structural KB, some researchers proposed to reason on implicit “KB”. 
Petroni \etal~[\citeyear{Petroni-EMNLP-2019}] analyzed the relational knowledge in a wide range of pretrained language models, and some follow-up work~\cite{Bouraoui-AAAI-2020,Jiang-TACL-2020} further demonstrated its effectiveness to answer “fill-in-the-blank” cloze statements. 
While most of existing work focused on  traditional structured KBs, a more general definition about KBs and flexible usage of KBs may help KBQA research topic show greater impact.

\end{itemize}

\section*{Acknowledgements}
This work is partially supported by the National Research Foundation, Singapore under its International Research Centres in Singapore Funding Initiative, the National Natural Science Foundation of China under Grant No. 61872369 and 61832017, Beijing Academy of Artificial Intelligence (BAAI) under Grant No. BAAI2020ZJ0301 and Beijing Outstanding Young Scientist Program under Grant No. BJJWZYJH012019100020098. Any opinions, findings and conclusions or recommendations expressed in this material are those of the author(s) and do not reflect the views of National Research Foundation, Singapore. Wayne Xin Zhao is the corresponding author.

\bibliographystyle{named}
\bibliography{survey}

\end{document}